\documentclass{article}

\usepackage{microtype}
\usepackage{graphicx}
\usepackage{subcaption}
\usepackage{booktabs}
\usepackage{svg}
\usepackage{hyperref}

\usepackage{fontawesome5}

\hypersetup{
    colorlinks=true,
    linkcolor=black,
    urlcolor=blue
}

\usepackage[preprint]{icml2026}

\usepackage{amsmath}
\usepackage{amssymb}
\usepackage{mathtools}
\usepackage{amsthm}
\usepackage{algorithm}
\usepackage{algorithmic}
\usepackage{multirow}
\usepackage{xcolor}
\usepackage{colortbl}
\usepackage{xspace}
\usepackage{enumitem}
\usepackage[capitalize,noabbrev]{cleveref}

\theoremstyle{plain}

\theoremstyle{definition}

\theoremstyle{remark}

\newcommand{\method}{SimpleTool\xspace}

\newcommand{\teaserfig}{%
  \includegraphics[width=0.95\textwidth]{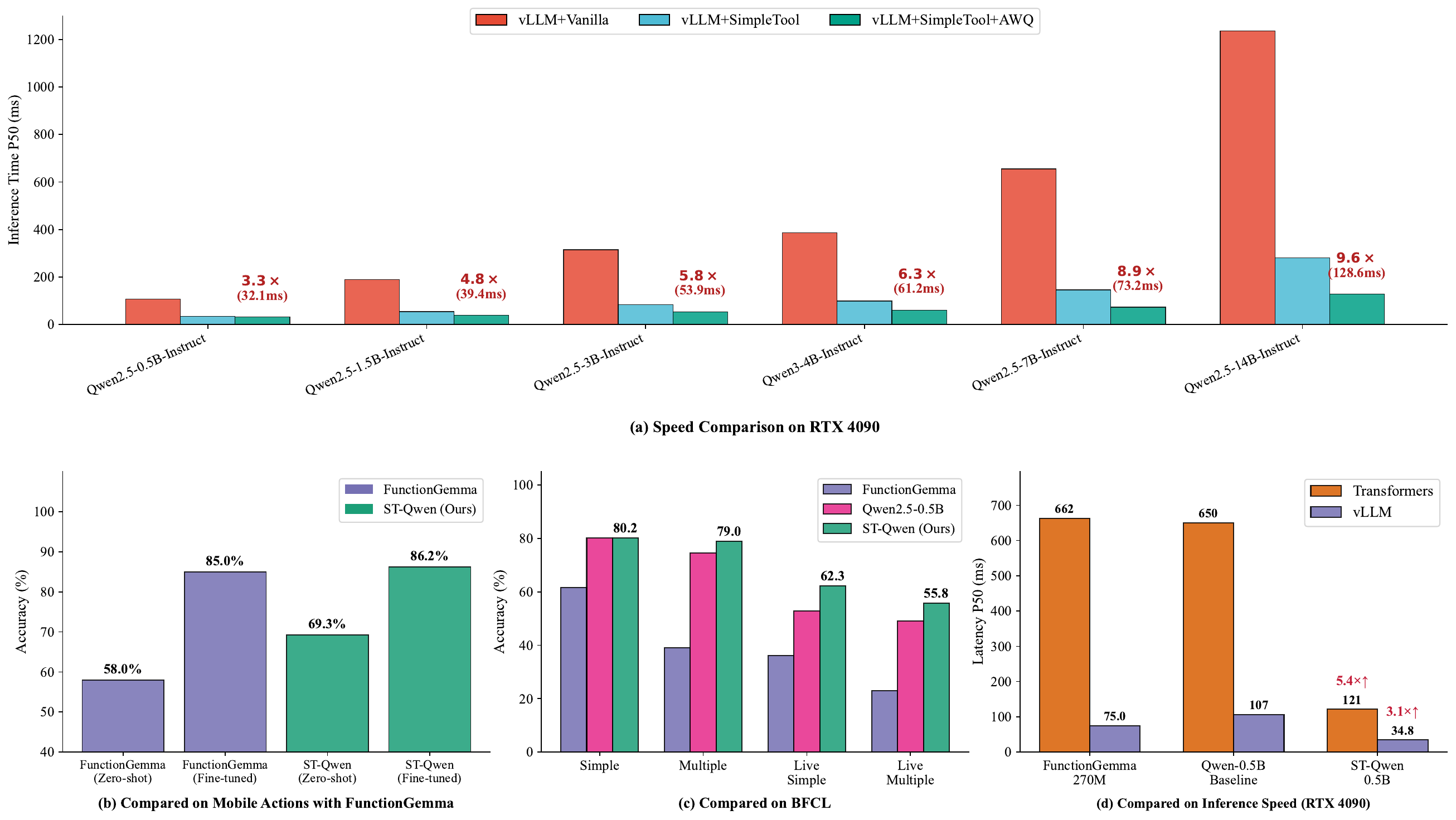}%
}

\hypersetup{
    colorlinks=true,
    linkcolor=black,
    urlcolor=blue,
    pdfinfo={
        Title={SimpleTool: Parallel Decoding for Real-Time LLM Function Calling},
        Author={Xiaoxin Shi, Jiaxin Wan, Linkang Dong, Wei Jiang, Yue Liu, Zengfeng Huang}
    }
}

\begin{document}

\twocolumn[
  \icmltitle{SimpleTool: Parallel Decoding for Real-Time LLM Function Calling}

  \icmlsetsymbol{equal}{*}

  \begin{icmlauthorlist}
    \icmlauthor{Xiaoxin Shi}{sjtu,sii}
    \icmlauthor{Jiaxin Wan}{sii}
    \icmlauthor{Linkang Dong}{sii}
    \icmlauthor{Wei Jiang}{sii}
    \icmlauthor{Yue Liu}{sii}
    \icmlauthor{Zengfeng Huang}{sii,fudan}
  \end{icmlauthorlist}

  \icmlaffiliation{sjtu}{Shanghai Jiao Tong University, Shanghai, China}
  \icmlaffiliation{sii}{Shanghai Innovation Institute, Shanghai, China}
  \icmlaffiliation{fudan}{Fudan University, Shanghai, China}

  \icmlcorrespondingauthor{Xiaoxin Shi}{cialtion737410@sjtu.edu.cn}
  \icmlcorrespondingauthor{Zengfeng Huang}{zfhuang@fudan.edu.cn}

  \vskip 0.2in
  \centering
  {\small
    \faGithub~\textbf{GitHub:} \href{https://github.com/HaxxorCialtion/SimpleTool}{https://github.com/HaxxorCialtion/SimpleTool}\\[4pt]
    \raisebox{-0.1em}{\includegraphics[height=1em]{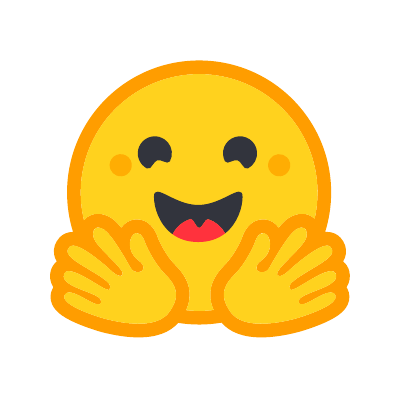}}~\textbf{Hugging Face:} \href{https://huggingface.co/Cialtion/SimpleTool}{https://huggingface.co/Cialtion/SimpleTool}\\[4pt]
    \raisebox{-0.1em}{\includegraphics[height=1em]{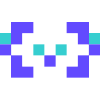}}~\textbf{ModelScope:} \href{https://www.modelscope.cn/models/cialtion/SimpleTool}{https://www.modelscope.cn/models/cialtion/SimpleTool}\\[4pt]
    \href{mailto:cialtion737410@sjtu.edu.cn}{cialtion737410@sjtu.edu.cn}\quad \href{mailto:huangzf@fudan.edu.cn}{huangzf@fudan.edu.cn}\\[4pt]
    $^1$Shanghai Jiao Tong University, Shanghai, China\\[2pt]
    $^2$Shanghai Innovation Institute, Shanghai, China\\[2pt]
    $^3$Fudan University, Shanghai, China
  }

  \vskip 0.3in
  \centering
  \teaserfig

  \vskip 0.5em
  \parbox{0.95\textwidth}{%
    \small \textbf{Figure 1.} Performance Overview of SimpleTool.
    \textbf{(a)} End-to-end inference latency comparison across various Qwen model sizes on an RTX 4090. Our method achieves up to a \textbf{9.6$\times$} speedup. 
    \textbf{(b)} Accuracy comparison against FunctionGemma on the Mobile Actions benchmark. 
    \textbf{(c)} Accuracy comparison on the BFCL benchmark among FunctionGemma, the Qwen2.5-0.5B baseline, and our ST-Qwen-0.5B. 
    \textbf{(d)} Inference latency comparison across different models and frameworks (Transformers vs. vLLM). 
    Latency measurements were conducted on an RTX 4090 using vLLM with prefix caching enabled for the BFCL-v3 benchmark evaluation. 
    Accuracy results for FunctionGemma are sourced from its official model card; all other results are obtained from our evaluation.%
  }
  \vskip 0.3in
]

\printAffiliationsAndNotice{Preprint. February 2026.}

\setcounter{figure}{1}

\begin{abstract}
LLM-based function calling enables intelligent agents to interact with external tools and environments,
yet autoregressive decoding imposes a fundamental latency bottleneck that limits real-time applications
such as embodied intelligence, game AI, and interactive avatars (e.g., 10\,Hz control frequency).
We observe that function calling differs fundamentally from free-form text generation:
structured outputs exhibit substantial token redundancy (delimiters, parameter names),
and arguments exhibit weak causal dependencies.
Crucially, these two properties must be exploited jointly to achieve real-time performance.
We present \method{}, which introduces special tokens that serve a dual role:
compressing low-entropy tokens (4--6$\times$ reduction) while acting as mode selectors 
that enable independent parallel generation of function name and arguments.
This synergistic design achieves 3--6$\times$ end-to-end speedup (up to 9.6$\times$) 
with only +8.2\% parallelization overhead.
Experiments on five benchmarks across Qwen-series models (0.5B--14B) demonstrate 
substantial speedup while maintaining competitive or improved accuracy.
On Mobile Actions, ST-Qwen-0.5B outperforms Google's FunctionGemma in both 
accuracy and latency consistency.
With quantization on consumer-grade GPU, \method{} achieves 
61.2ms P50 latency, enabling 16\,Hz real-time control at 4B model scale,
bridging the gap between LLM function calling and latency-critical real-world deployment.
\end{abstract}

\section{Introduction}
\label{sec:intro}

Large Language Models (LLMs) have enabled function calling---the mechanism 
by which LLMs interact with external tools and environments 
\cite{brown2020language,o2024open,schick2023toolformer}. 
However, a fundamental limitation persists: autoregressive decoding 
imposes a per-token latency floor that scales linearly with output length 
\cite{leviathan2023fast,cai2024medusa}. 
While streaming masks this latency for conversational applications, 
function calling requires \textbf{complete and valid output before execution}---a 
partial function call is semantically meaningless. 
This creates an end-to-end latency bottleneck that streaming cannot alleviate.

\subsection{The Real-Time Gap}

This bottleneck creates a significant gap between LLM capabilities and 
real-world deployment requirements (Figure~1a).

\textbf{Embodied AI \& Game AI.} 
Real-time control demands 5--30\,Hz response frequencies. 
Vision-language-action models like OpenVLA \cite{kim2025openvla} achieve 
only 6\,Hz at 166\,ms latency \cite{yu2025survey}---far below 
the sub-100\,ms threshold for responsive control.
Game AI systems like Lumine \cite{tan2025lumine} achieve 5\,Hz 
only through extensive datacenter infrastructure.

\textbf{Edge Deployment.} 
Mobile agents \cite{yang2025qwen3,wang2024mobile} rely on 
cloud APIs with 500\,ms--2\,s latency per action.
Google's FunctionGemma \cite{functiongemma2025}---a 270M model 
for edge function calling---trades capacity for speed, 
underscoring the need for efficient function calling methods.

\subsection{Rethinking Acceleration for Function Calling}

Function calling end-to-end latency determines the upper bound of 
LLM-based system capabilities, making acceleration critical. However,
existing acceleration methods are not designed for this setting.

\textbf{Speculative and Parallel Decoding.} 
Methods like Medusa \cite{cai2024medusa}, EAGLE \cite{li2024eagle}, 
and draft-model approaches \cite{leviathan2023fast} 
accelerate per-token generation through speculation and verification.
Though effective for long-form generation, their acceleration benefits are 
diminished by associated overhead in short-output scenarios such as function calls.
We take a more fundamental approach: exploiting the memory-bandwidth bottleneck 
and idle compute capacity during decoding, combined with the inherent weak causal 
dependencies in function calling, to enable concurrent decoding of function name 
and arguments---achieving aggressive end-to-end acceleration.
Moreover, these approaches are \textbf{orthogonal} to ours and can be combined:
we validate this in Appendix~\ref{app:speculative}, 
showing that speculative decoding can achieve up to 3$\times$ additional 
forward pass reduction with $>$93\% token acceptance rate when 
applied to \method{} models.

\textbf{Constrained Decoding.} 
Grammar-guided frameworks like SGLang \cite{zheng2024sglang}, 
Outlines \cite{willard2023efficient}, and XGrammar \cite{dong2024xgrammar} 
ensure output validity and reduce some redundant tokens. Their methods can also 
reduce end-to-end latency in function calls. 
However, they remain fundamentally autoregressive, 
generating tokens sequentially without addressing the core bottleneck: 
token-by-token decoding.

\textbf{Quantization \& Compression.} 
Model compression techniques \cite{dettmers2022gpt3,frantar2023gptq,lin2025awq} 
provide general acceleration applicable to any task.
Our method combines naturally with these approaches 
(Table~\ref{tab:speedup}).

These methods make significant improvements in various aspects.
However, they still cannot enable LLMs to make function calls in real time.
To achieve this, we advocate a more radical approach 
that not only leverages parallel decoding but also drastically reduces 
the number of tokens to be decoded. Furthermore, these two aspects are 
mutually reinforcing.

\subsection{Key Insight and Our Approach}

We observe that function calls generated by LLMs---typically in JSON or 
Python API format---exhibit two key properties: 
weak causal dependencies among arguments and substantial token redundancy.
Based on this insight, we propose \textbf{\method{}}, which achieves 
real-time function calling through two synergistic mechanisms 
(Figure~\ref{fig:overview}):

\begin{figure}[t]
    \centering
    \includegraphics[width=\columnwidth]{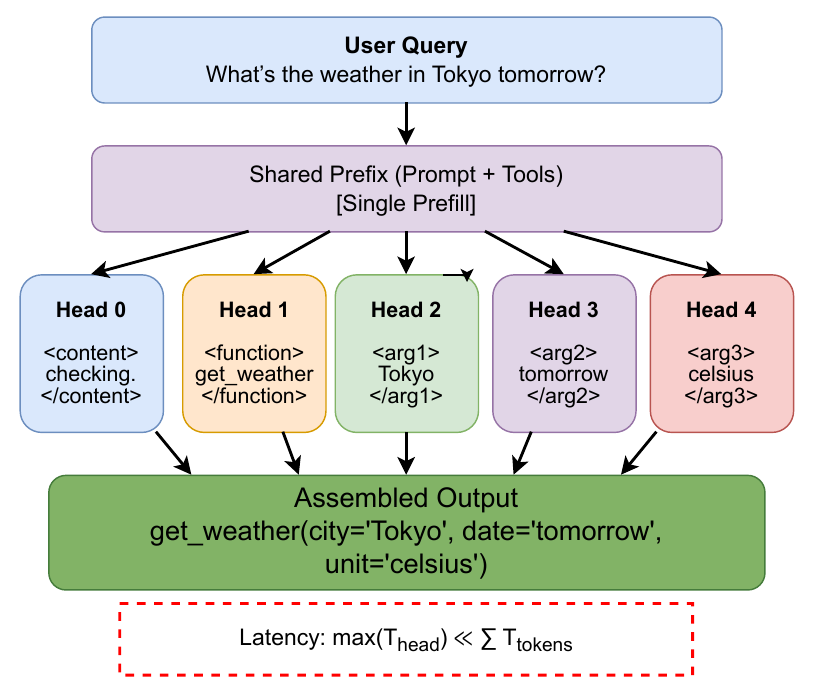}
    \caption{Overview of \method{}. Given an input prompt with tool 
    definitions, parallel heads generate function name and arguments 
    independently while sharing the prefix KV cache.}
    \label{fig:overview}
\end{figure}

\begin{figure}[t]
    \centering
    \includegraphics[width=\columnwidth]{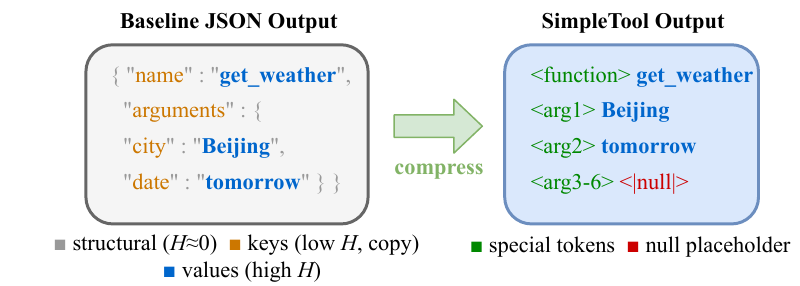}
    \vspace{-0.5em}
    \caption{Token compression illustration. Baseline structured output 
    contains $\sim$30 tokens spanning three entropy levels. \method{} 
    compresses low-entropy tokens into special markers, 
    retaining only high-entropy values for generation.}
    \label{fig:compression}
    \vspace{-0.5em}
\end{figure}

\textbf{Special Tokens.} 
We introduce special tokens specifically designed for function calling, 
which are injected into the vocabulary and can effectively compress redundant tokens. 
Figure~\ref{fig:compression} illustrates this mechanism. 
We demonstrate the effectiveness of special tokens in Table~\ref{tab:token_compression}:
\textbf{4--6$\times$ reduction} compared with vanilla JSON output 
on samples where both models produce correct outputs. 
Moreover, these special tokens facilitate parallel decoding as described below.

\textbf{Parallel Decoding.} 
Special tokens further weaken the causal dependencies inherent in function calling, 
allowing us to treat the function name and its arguments as independent streams. 
To achieve ultimate speedup, we decode the function name and arguments in parallel. 
These parallel streams share the same input prefix and differ only in the 
appended special token, requiring only a single prefill operation with full 
reuse of the prefix KV cache. This drastically reduces memory overhead and 
enables full utilization of idle compute capacity. 
Consequently, negligible overhead is observed in the 
time per output token (TPOT) test presented in Appendix~\ref{app:batch_scaling}.

\subsection{Results Overview and Contributions}

We conduct comprehensive evaluation on Qwen-series models 
ranging from 0.5B to 14B parameters across \textbf{five benchmarks}: 
BFCL-v3 \cite{patil2025the}, Mobile Actions \cite{functiongemma2025}, 
SealTools \cite{wu2024seal}, OpenFunction \cite{patil2024gorilla}, 
and ToolAlpaca \cite{tang2023toolalpaca}, 
covering diverse function calling scenarios from API invocation to 
mobile device control.

Results demonstrate that \method{} achieves \textbf{3--6$\times$ speedup} 
with competitive or improved accuracy across all model sizes and benchmarks. 
Combined with quantization, our method enables sub-100ms function calling latency 
with 4B-scale models on consumer-grade GPUs, reaching real-time control frequencies 
previously infeasible with autoregressive decoding.
We additionally compare against Google's recent model, FunctionGemma \cite{functiongemma2025}, 
on the Mobile Actions benchmark. In both zero-shot and fine-tuned settings, 
our smallest model, ST-Qwen2.5-0.5B, demonstrates advantages in both 
accuracy and latency for edge deployment scenarios.

\textbf{Contributions.}
\begin{enumerate}[label=(\arabic*)]
    \item We propose \textbf{\method{}}, a parallel decoding framework that 
    enables LLMs to call functions in real time, making it possible for LLM-based 
    methods to control real-time interactive systems.
    
    \item We design mutually reinforcing special tokens and a parallel decoding 
    architecture that exploits idle compute capacity, conducting an in-depth 
    exploration of redundant token compression and parallel decoding for function calling.
    
    \item We train multiple \method{} models across the Qwen series (0.5B--14B) 
    and evaluate them on five benchmarks, demonstrating substantial speedup with 
    competitive accuracy.
    
    \item We compare against Google's FunctionGemma on Mobile Actions in both 
    zero-shot and fine-tuned settings, showing more effective edge deployment capability.
\end{enumerate}
\section{Methodology}
\label{sec:method}

Building on the insight that function calling exhibits structural redundancy 
and weak causal dependencies (Section~\ref{sec:intro}), this section presents 
the technical realization of \method{}. 
The key challenge is that these two properties must be exploited \textit{jointly}: 
compressing redundant tokens alone does not break sequential decoding, 
while parallelizing without compression leaves substantial latency unreduced.

We describe how special tokens and parallel decoding work synergistically: 
special tokens not only compress output length (4--6$\times$ reduction) but also 
serve as \textbf{mode selectors} that enable independent generation of function 
name and arguments (\Cref{sec:parallel_decoding}). 
The training strategy then ensures the model learns to leverage both mechanisms 
effectively (\Cref{sec:training strategy}).

\subsection{Special Tokens Enable Parallel Decoding}
\label{sec:parallel_decoding}

\subsubsection{From Redundancy to Compression}

Intuitively, conventional function call outputs contain three token categories with 
distinct entropy characteristics. (Figure~\ref{fig:compression}):

\begin{itemize}
    \setlength{\itemsep}{0pt}
    \setlength{\parskip}{0pt}
    \item \textbf{Structural tokens} (\texttt{\{}, \texttt{\}}, 
    \texttt{:}, \texttt{,}): Fully predictable given the output format, near-zero entropy.
    \item \textbf{Key tokens} (``name'', ``arguments'', parameter names): 
    Copied from tool definitions, low entropy.
    \item \textbf{Value tokens} (function names, argument values): 
    Require actual LLM reasoning, high entropy.
\end{itemize}

We introduce special tokens (\texttt{<function>}, \texttt{<arg1>}--\texttt{<arg6>} e.g.) 
that absorb all low-entropy information, achieving 4--6$\times$ compression 
(Table~\ref{tab:token_compression}). Crucially, these tokens serve a dual purpose: 
beyond compression, each token acts as a \textbf{mode selector}, signaling which component 
(function name, specific argument or even content) to generate. This dual role---compression \textit{and} mode selection---is 
what enables parallel decoding.

\begin{table}[t]
\centering
\caption{Token compression ratio across model sizes. 
CR = Baseline tokens / \method{} bottleneck-head tokens. 
Statistics computed on samples where both models produce correct outputs. 
Per-benchmark breakdown in Appendix~\ref{app:compression_details}.}
\label{tab:token_compression}
\small
\begin{tabular}{lccc}
\toprule
\textbf{Model} & \textbf{CR (Mean)} & \textbf{CR (P50)} & \textbf{CR (P90)} \\
\midrule
Qwen2.5-0.5B & 5.06$\times$ & 6.00$\times$ & 4.06$\times$ \\
Qwen2.5-1.5B & 4.32$\times$ & 4.71$\times$ & 3.20$\times$ \\
Qwen2.5-3B & 4.38$\times$ & 4.71$\times$ & 3.20$\times$ \\
Qwen3-4B & 4.32$\times$ & 4.86$\times$ & 3.42$\times$ \\
Qwen2.5-7B & 5.35$\times$ & 5.00$\times$ & 3.40$\times$ \\
Qwen2.5-14B & 4.51$\times$ & 4.86$\times$ & 3.47$\times$ \\
\midrule
\textbf{Average} & \textbf{4.66$\times$} & \textbf{5.02$\times$} & \textbf{3.46$\times$} \\
\bottomrule
\end{tabular}
\end{table}

\subsubsection{From Weak Dependencies to Parallel Decoding}

With special tokens serving as mode selectors, we can exploit the weak causal 
dependencies among function arguments: since each argument can be inferred 
largely independently from the input context, we decode them as 
\textbf{parallel streams} rather than sequentially.

\textbf{Architecture.} 
As illustrated in Figure~\ref{fig:overview}, \method{} decouples 
function calling into $H$ independent decoding streams 
(1 for function name, up to 6 for arguments, more if needed in training), 
each initialized by appending a distinct special token 
(\texttt{<function>}, \texttt{<arg1>}, ..., \texttt{<arg6>}) to the shared input prefix.  
We also reserve a content head (\texttt{<content>}) for chatting, which
can be optionally used alongside function calls.
All streams share the same prefix KV cache; 
only the head-specific special token differs.
This design directly leverages the dual role of special tokens: 
they compress the output \textit{and} specify which component to generate.

\textbf{Latency Analysis.} 
For conventional autoregressive decoding, end-to-end latency is:
\begin{equation}
    T_{\text{baseline}} = T_p + N \cdot T_d
\end{equation}
where $T_p$ is prefill time, $N$ is total output tokens, 
and $T_d$ is per-token decode latency. 
With \method{}, the latency becomes:
\begin{equation}
    T_{\text{ours}} \approx T_p + \max_{i \in \{1,...,H\}}(N_i) \cdot T_d
    \label{eq:latency}
\end{equation}
where $H$ is the number of parallel decoding heads, 
$N_i$ is the token count for head $i$, 
and some insignificant overhead may exist.
End-to-end latency is thus dominated by the \textbf{longest single head}, 
not the sum of all heads.

\subsubsection{Parallel Decoding as a Near-Free Operation}

One might expect $H$ parallel decoding streams to incur 
proportional computational overhead. 
We argue---and empirically validate---that this cost is negligible 
due to the \textbf{memory-bandwidth-bound} nature of autoregressive decoding.

During the decode phase, each token generation requires loading 
the full model weights from GPU memory, while the actual computation 
(a single matrix-vector product per layer) utilizes only a fraction 
of available FLOPS. This creates substantial idle compute capacity.
Batching multiple sequences---in our case, parallel heads within 
the same request---increases arithmetic intensity without proportionally 
increasing memory traffic, effectively utilizing otherwise-idle compute.

We validate this through batch scaling experiments (shown in Appendix~\ref{app:batch_scaling}). 
Across six model configurations from 0.5B to 14B parameters 
with the same prefix and one different token, 
8-head parallel decoding achieves \textbf{93.0\% average TPOT efficiency},
 with efficiency only dropping markedly at extremely large batch sizes, 
demonstrating that \method{}'s parallel heads operate well within 
the memory-bound bottleneck and idle compute capacity.

\textbf{KV Cache Efficiency.} 
All parallel heads share the input prefix, enabling full KV cache reuse. 
For an input of $L$ tokens and $H$ heads each appending a single special token, 
the KV cache overhead is only $H$ tokens beyond the shared prefix---negligible 
compared to typical prompt lengths.
In high-frequency control scenarios where the tool schema remains fixed, 
this shared prefix can be persistently cached, further amortizing prefill cost 
(see Section~\ref{sec:exp_mobile} for empirical validation).

\subsection{Training for Parallel Generation}
\label{sec:training strategy}

The synergistic design described above requires the model to learn 
a fundamentally different output behavior: generating function name 
and arguments as \textbf{independent streams} conditioned on distinct 
special tokens, while maintaining reasoning capability.

\subsubsection{Special Token Design}

We introduce 17 special tokens organized into four functional groups:

\noindent
\textbf{Content head}: \texttt{<content>}, \texttt{</content>} for natural language responses. \\
\textbf{Function head}: \texttt{<function>}, \texttt{</function>} for function name generation. \\
\textbf{Argument heads}: \texttt{<arg$k$>}, \texttt{</arg$k$>} for $k \in \{1,...,6\}$, each selecting a positional argument. \\
\textbf{Placeholder}: \texttt{<|null|>} for unused argument slots, enabling early termination of inactive heads.

Each special token encodes all structural information 
(delimiters, parameter names, positional ordering) into its semantics, 
eliminating the need to generate them explicitly while simultaneously 
specifying the generation mode.

\subsubsection{Training Challenges}

Achieving high-quality parallel decoding is non-trivial. 
Unlike standard fine-tuning where the model learns a single output distribution, 
\method{} requires the model to learn \textbf{eight distinct output modes} 
that share the same input but produce semantically different outputs.
This poses two key challenges:

\textbf{Capacity Requirements.} 
Learning multiple output modes demands sufficient model plasticity. 
We find that standard LoRA configurations underperform for this task; 
significantly larger adapter capacity is required to capture the 
diverse output patterns without interference between heads 
(see ablation in Section~\ref{sec:ablation_rank}).
Moreover, the introduction of special tokens necessitates 
unfreezing the embedding layer---a departure from standard 
parameter-efficient fine-tuning that requires careful optimization 
to balance learning new token representations against preserving 
pretrained knowledge.

\textbf{Data Distribution Balance.} 
Existing function calling datasets exhibit skewed argument count 
distributions, with most samples concentrated around 2--3 arguments. 
For \method{}'s multi-head architecture, 
underrepresented argument counts (0, 5, 6 arguments) lead to 
poorly trained heads that fail to generalize.
We address this through targeted data augmentation using synthetic 
samples that balance argument count distribution 
(see ablation in Section~\ref{sec:ablation_data}).

\subsubsection{Training Pipeline}

We adopt parameter-efficient fine-tuning with LoRA \cite{hu2022lora}, 
targeting the MLP layers to maximize the capacity for learning 
head-specific output patterns. Training samples are formatted as 
parallel sequences sharing identical prefixes but differing in 
the appended special token and corresponding target output.
The loss is computed independently for each head with 
head-specific weighting to address output length imbalance.
Hyperparameter details are provided in Appendix~\ref{app:hyperparameters}.

\section{Experiments}
\label{sec:experiments}

We evaluate \method{} from four perspectives: 
(1) \textbf{accuracy} on multiple benchmarks (Section~\ref{sec:exp_accuracy}), 
(2) \textbf{inference speedup} (Section~\ref{sec:exp_speedup}),
(3) \textbf{domain adaptation} comparing with FunctionGemma 
(Section~\ref{sec:exp_mobile}), and 
(4) \textbf{ablation studies} (Section~\ref{sec:exp_ablation}).

\subsection{Experimental Setup}
\label{sec:exp_setup}

\textbf{Models.} We evaluate on Qwen2.5-Instruct series 
(0.5B--14B) \cite{qwen2.5} and Qwen3-4B-Instruct \cite{yang2025qwen3}. 
All models are fine-tuned using LoRA \cite{hu2022lora} with a relatively 
large rank (we explore this setting in 
Section~\ref{sec:ablation_rank}). We set the number of parallel argument heads to 6, which already covers 
\textbf{95.2\%} of function calls across all evaluation benchmarks 
(see Appendix~\ref{app:param_stats} for detailed statistics).

\textbf{Benchmarks.} We evaluate on five benchmarks covering diverse 
function calling scenarios: 
BFCL-v3 \cite{patil2024gorilla} (single-turn subsets), 
Mobile Actions \cite{functiongemma2025}, 
SealTools \cite{wu2024seal}, 
OpenFunction-v1 \cite{patil2024gorilla}, and 
ToolAlpaca \cite{tang2023toolalpaca}. The last three benchmarks are 
combined into an ``Others'' group for aggregate evaluation.

\textbf{Metrics.} Overall Accuracy (complete correctness including 
function name and all arguments) and Function Accuracy (function 
name selection only).

\textbf{Hardware.} Inference latency is measured on RTX 4090 and H100 GPUs.

\subsection{Accuracy Evaluation}
\label{sec:exp_accuracy}

\begin{figure*}[t]
    \centering
    \includegraphics[width=0.85\textwidth]{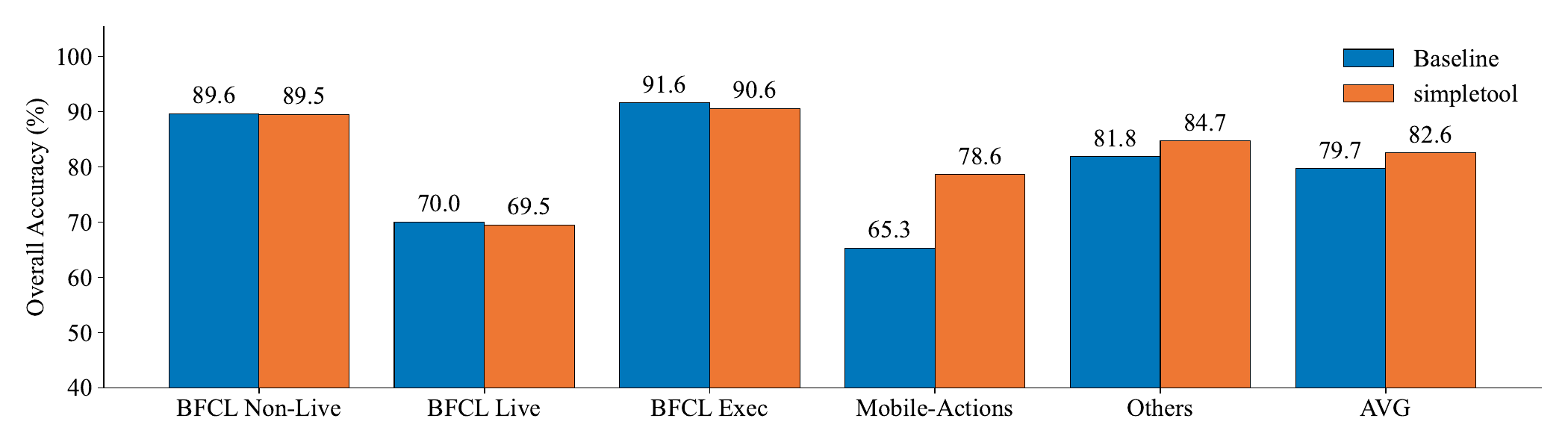}
    \vspace{-0.5em}
    \caption{Average accuracy comparison between Baseline and \method{} 
    across benchmark groups (macro average). 
    BFCL-v3 is evaluated on single-turn subsets; 
    Mobile Actions parallel calls are converted to multi-turn format; 
    ``Others'' combines SealTools, OpenFunc, and ToolAlpaca.}
    \label{fig:accuracy}
    \vspace{-0.5em}
\end{figure*}

\begin{table*}[t]
  \centering
  \caption{Performance comparison on five benchmarks. 
    Each cell shows overall (function) accuracy, e.g., 78.3 (98.3). 
    Speedup ratio after model name indicates inference acceleration 
    on RTX 4090 with vLLM (P50 latency).
    \textbf{Bold} = best, \underline{underline} = second best 
    within each column (applied separately to overall and function accuracy).}
    \label{tab:main_results}
  \small
  \setlength{\tabcolsep}{6pt}
  \begin{tabular}{lcccccc}
    \toprule
    \textbf{Model} & \textbf{BFCL Non-Live} & \textbf{BFCL Live} & \textbf{BFCL Exec} & \textbf{Mobile Actions} & \textbf{Others} & \textbf{AVG} \\
    \midrule
    \multicolumn{7}{l}{\textit{Baseline Models}} \\
    Qwen2.5-0.5B (1$\times$) & 78.3 (98.3) & 49.9 (86.4) & 88.6 (96.6) & 56.7 (74.4) & 80.9 (95.2) & 70.9 (90.2) \\
    Qwen2.5-1.5B (1$\times$) & 86.3 (98.3) & 67.4 (90.4) & 91.9 (98.0) & 65.5 (74.9) & 82.2 (95.3) & 78.7 (91.4) \\
    Qwen2.5-3B (1$\times$) & 92.5 (\underline{99.7}) & 73.0 (94.3) & 86.6 (96.0) & 66.2 (75.0) & 84.3 (95.9) & 80.5 (92.2) \\
    Qwen3-4B (1$\times$) & \underline{93.7} (99.3) & \textbf{77.9} (94.5) & \textbf{94.6} (97.3) & 68.0 (75.0) & 82.0 (91.8) & 83.2 (91.6) \\
    Qwen2.5-7B (1$\times$) & 93.0 (\underline{99.7}) & 75.4 (95.2) & \underline{94.0} (\underline{99.3}) & 68.4 (77.2) & 83.4 (94.9) & 82.8 (93.3) \\
    Qwen2.5-14B (1$\times$) & \textbf{94.0} (99.3) & \underline{76.6} (95.9) & \underline{94.0} (98.7) & 67.0 (75.0) & 78.3 (89.5) & 82.0 (91.7) \\
    \midrule
    \multicolumn{7}{l}{\textit{SimpleTool (Ours)}} \\
    ST-Qwen2.5-0.5B (3.1$\times$) & 79.8 (\textbf{99.8}) & 57.2 (91.6) & 87.9 (\textbf{100.0}) & 69.3 (99.3) & 82.2 (99.1) & 75.3 (98.0) \\
    ST-Qwen2.5-1.5B (3.5$\times$) & 88.8 (\textbf{99.8}) & 63.6 (94.3) & 90.6 (\textbf{100.0}) & 72.8 (\textbf{100.0}) & 80.2 (99.3) & 79.2 (98.7) \\
    ST-Qwen2.5-3B (3.8$\times$) & 90.3 (\textbf{99.8}) & 68.0 (95.4) & 91.3 (\textbf{100.0}) & 81.4 (\underline{99.9}) & 85.5 (\underline{99.5}) & 83.3 (98.9) \\
    ST-Qwen3-4B (3.9$\times$) & 92.5 (99.5) & 76.4 (96.3) & 89.9 (\textbf{100.0}) & \textbf{84.5} (\underline{99.9}) & \underline{86.6} (\textbf{99.7}) & \underline{86.0} (99.1) \\
    ST-Qwen2.5-7B (4.5$\times$) & 93.5 (\textbf{99.8}) & 75.8 (\underline{96.7}) & 92.6 (\textbf{100.0}) & \underline{83.2} (\textbf{100.0}) & 86.2 (\underline{99.5}) & \textbf{86.3} (\underline{99.2}) \\
    ST-Qwen2.5-14B (4.4$\times$) & 92.2 (\underline{99.7}) & 75.9 (\textbf{97.1}) & 91.3 (\textbf{100.0}) & 80.5 (\textbf{100.0}) & \textbf{87.4} (\textbf{99.7}) & 85.5 (\textbf{99.3}) \\
    \bottomrule
  \end{tabular}
\end{table*}

Table~\ref{tab:main_results} and Figure~\ref{fig:accuracy} present 
comprehensive accuracy evaluation across model scales and benchmarks.

\textbf{Consistent improvement across model scales.} 
All model sizes show overall accuracy gains ranging from +0.5\% (1.5B) 
to +3.5\% (7B), with an average improvement of +2.9\%. 
This consistency suggests that models have learned these new output modes effectively.

\textbf{Substantial function accuracy gains.} 
Average function accuracy increases +7.1\% over baselines, 
with several configurations achieving near-perfect scores. 
We attribute this to special tokens acting as mode selectors: 
by explicitly conditioning each head on its target component, 
the model avoids ambiguity in output structure. 
Additionally, the compressed output format means fewer tokens 
and thus fewer opportunities for errors---particularly important 
in function calling where a single incorrect token invalidates 
the entire output.

\textbf{Strongest gains on Mobile Actions.} 
Improvements range from +7.3\% to +13.3\% across model sizes. 
Notably, this benchmark was released in December 2025 \cite{functiongemma2025}, 
after both baseline models and our training data were finalized, 
demonstrating strong generalization to unseen domains.

\subsection{Speedup Evaluation}
\label{sec:exp_speedup}

\begin{figure*}[t]
    \centering
    \includegraphics[width=0.95\textwidth]{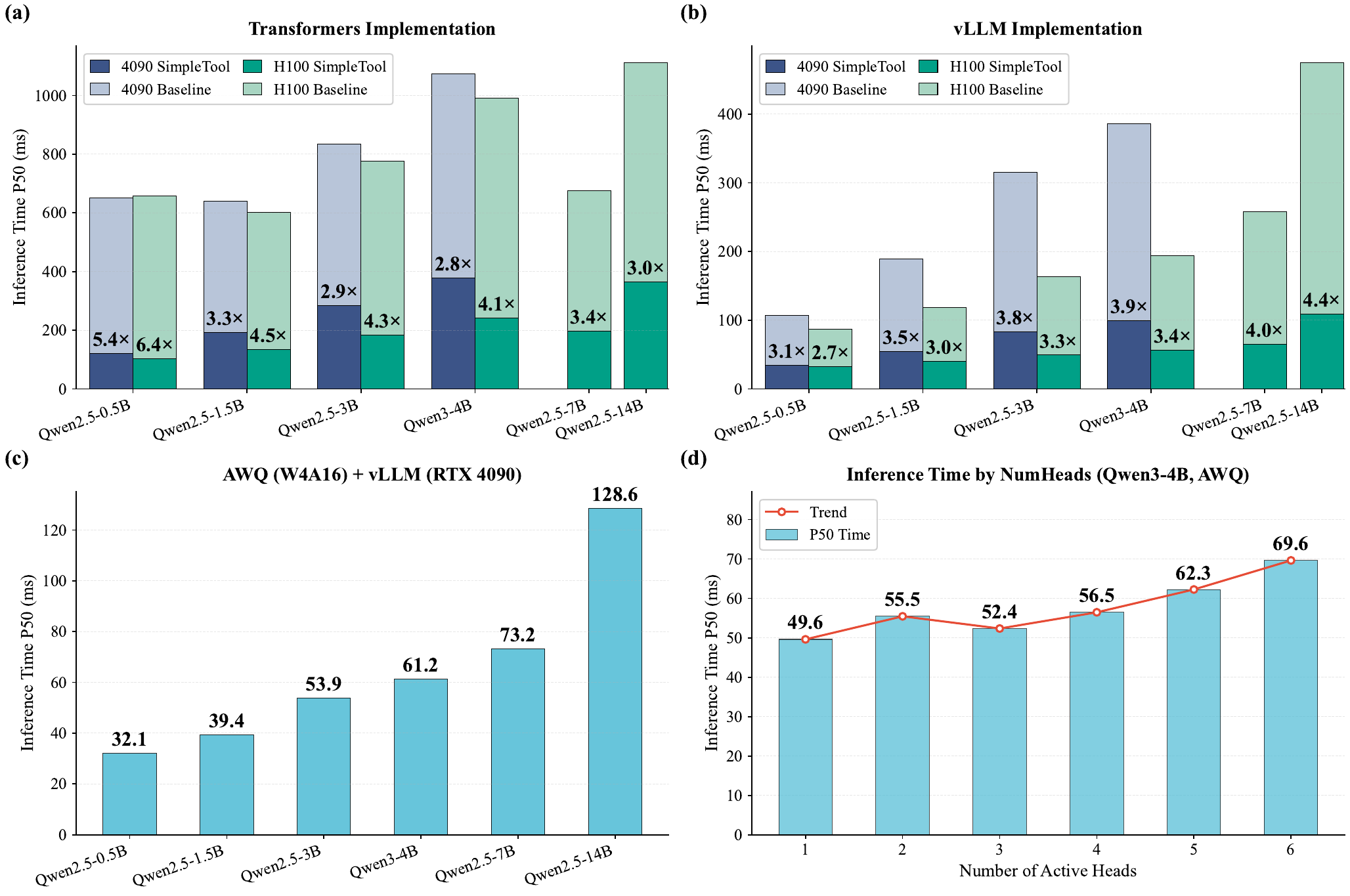}
    \vspace{-0.5em}
    \caption{\textbf{Inference speedup evaluation.}
        (a)--(b) Stacked bars show baseline time partitioned into 
        SimpleTool time (dark) and time saved (light); 
        speedup ratios labeled at boundary.
        (a) Transformers backend on RTX 4090 (blue) and H100 (green).
        (b) vLLM backend with prefix caching.
        (c) Absolute latency with AWQ quantization on RTX 4090.
        (d) Latency scaling with number of active heads (Qwen3-4B, AWQ in 4090).
        All measurements on BFCL-v3 single-turn cases.}
    \label{fig:speedup}
    \vspace{-0.5em}
\end{figure*}

Figure~\ref{fig:speedup} presents speedup evaluation in different settings,
demonstrating the effectiveness of \method{}'s acceleration strategies.

\textbf{Basic Parallelization} (Figure~\ref{fig:speedup}a): 
With the Transformers backend, \method{} achieves 
2.8--5.4$\times$ speedup on RTX 4090 and 3.0--6.4$\times$ on H100. 
Smaller models benefit more due to their decode-bound characteristics.

\textbf{Prefix Caching} (Figure~\ref{fig:speedup}b): 
vLLM's KV cache sharing across parallel streams yields 
2.7--4.4$\times$ speedup, with larger models showing improved 
ratios as prefix caching amortizes shared computation more effectively.

\textbf{Quantization} (Figure~\ref{fig:speedup}c): 
AWQ 4-bit quantization achieves the lowest absolute latency. 
Qwen3-4B reaches \textbf{61.2ms} P50, corresponding to 
16\,Hz control frequency---exceeding the 10\,Hz threshold 
typically required for real-time applications.

\textbf{Parallelization Overhead} (Figure~\ref{fig:speedup}d): 
As active heads increase from 2 to 8, speedup remains stable, 
confirming that parallelization overhead is negligible---consistent 
with our analysis that parallel streams utilize otherwise-idle 
compute capacity within the memory-bandwidth-bound decode phase. 
Note that Figure~\ref{fig:speedup}d reflects end-to-end measurements 
where varying output lengths across heads introduce additional variance; 
controlled experiments isolating batch size effects are provided in 
Appendix~\ref{app:batch_scaling}.

\begin{table}[ht]
\centering
\small
\caption{Speedup and latency summary on RTX 4090.}
\label{tab:speedup}
\begin{tabular*}{\linewidth}{@{\extracolsep{\fill}}lccc@{}}
\toprule
\textbf{Model} & \textbf{Trans.} & \textbf{vLLM} & \textbf{AWQ (ms)} \\
\midrule
Qwen2.5-0.5B & 5.35$\times$ & 3.07$\times$ & 32.1 \\
Qwen2.5-1.5B & 3.32$\times$ & 3.46$\times$ & 39.4 \\
Qwen2.5-3B & 2.93$\times$ & 3.78$\times$ & 53.9 \\
Qwen3-4B & 2.83$\times$ & 3.88$\times$ & 61.2 \\
Qwen2.5-7B & -- & -- & 89.3 \\
Qwen2.5-14B & -- & -- & 142.5 \\
\bottomrule
\end{tabular*}
\end{table}

Table~\ref{tab:speedup} summarizes results across the acceleration hierarchy. 
Models up to 7B achieve sub-100ms latency on consumer-grade RTX 4090, 
enabling real-time deployment without datacenter infrastructure.

\subsection{Case Study: Comparison with FunctionGemma}
\label{sec:exp_mobile}

We compare against FunctionGemma \cite{functiongemma2025}, Google's 
270M model released in December 2025 specifically for edge function calling, 
to evaluate whether \method{} can compete with industry solutions 
designed for on-device deployment.

\textbf{Setup.} We evaluate on the Mobile Actions dataset released 
alongside FunctionGemma, containing $\sim$9,600 samples covering 
7 smartphone operations. This benchmark complements BFCL-v3 by 
focusing on mobile device control---a scenario closely aligned with 
our target real-time applications.
For domain adaptation, we fine-tune ST-Qwen-0.5B on the official 
training split using LoRA ($r=64$), requiring less than 3 H100-hours. 
Latency is measured on the full test split (1,283 samples) 
using vLLM with prefix caching.

\begin{table}[t]
\centering
\small
\caption{Accuracy on Mobile Actions.}
\label{tab:mobile_accuracy}
\begin{tabular*}{\linewidth}{@{\extracolsep{\fill}}lcc@{}}
\toprule
\textbf{Model} & \textbf{Zero-shot} & \textbf{Fine-tuned} \\
\midrule
Qwen3-4B-Instruct & 83.7 & -- \\
FunctionGemma (270M) & 58.0 & 85.0 \\
ST-Qwen-0.5B (Ours) & \textbf{69.3} & \textbf{86.2} \\
\midrule
$\Delta$ vs FunctionGemma & +11.3 & +1.2 \\
\bottomrule
\end{tabular*}
\end{table}

\begin{table}[t]
\centering
\small
\caption{Latency on Mobile Actions (vLLM, RTX 4090).}
\label{tab:mobile_speed}
\begin{tabular*}{\linewidth}{@{\extracolsep{\fill}}lccc@{}}
\toprule
\textbf{Model} & \textbf{P50 (ms)} & \textbf{P90 (ms)} & \textbf{Speedup} \\
\midrule
Qwen3-4B (Baseline) & 468.5 & 667.0 & 1.0$\times$ \\
Qwen2.5-0.5B (Baseline) & 110.1 & 154.0 & 4.3$\times$ \\
FunctionGemma (270M) & 61.1 & 139.5 & 7.7$\times$ \\
\midrule
\textbf{ST-Qwen-0.5B (Ours)} & \textbf{51.0} & \textbf{74.5} & \textbf{9.2$\times$} \\
\bottomrule
\end{tabular*}
\end{table}

\textbf{Results} (Tables~\ref{tab:mobile_accuracy}--\ref{tab:mobile_speed}):
ST-Qwen-0.5B achieves 69.3\% zero-shot (+11.3\% vs FunctionGemma) and 
86.2\% after fine-tuning, surpassing both FunctionGemma (85.0\%) and 
the 8$\times$ larger Qwen3-4B (83.7\%). Despite 1.8$\times$ more parameters, 
ST-Qwen-0.5B achieves faster inference (51.0ms vs 61.1ms P50) with 
substantially better tail latency (74.5ms vs 139.5ms P90)---critical 
for real-time applications where worst-case latency determines reliability.

\subsection{Ablation Studies}
\label{sec:exp_ablation}

We conduct ablation studies on Qwen3-4B evaluated on BFCL-v3 (Table~\ref{tab:ablation}).

\textbf{LoRA Rank.}
\label{sec:ablation_rank}
Performance improves with larger ranks, reaching 87.0\% at rank 1024, 
suggesting that learning eight distinct output modes benefits from 
increased adapter capacity. We adopt rank 512 as default.

\begin{table}[h]
\centering
\small
\caption{Ablation studies on BFCL-v3 (Qwen3-4B).}
\label{tab:ablation}
\begin{tabular}{lc|lc}
\toprule
\textbf{LoRA Rank} & \textbf{Acc. (\%)} & \textbf{Training Data} & \textbf{Acc. (\%)} \\
\midrule
64 & 85.5 & xLAM only (public) & 85.3 \\
256 & 86.4 & xLAM + Synthetic & \textbf{86.3} \\
512 & 85.3 & & \\
1024 & \textbf{87.0} & & \\
\bottomrule
\end{tabular}
\end{table}

\textbf{Training Data.}
\label{sec:ablation_data}
Using only public xLAM data \cite{zhang2025xlam}, \method{} achieves 
85.3\%---demonstrating strong performance without proprietary data. 
Synthetic augmentation via Qwen3-235B yields +1.0\% improvement.

\section{Limitations}
\label{sec:limitations}

\noindent\textbf{Argument Independence.} 
\method{} assumes function arguments can be generated independently---valid 
for well-designed APIs with unordered key-value parameters. When semantic 
dependencies exist (e.g., \texttt{file\_path} and \texttt{line\_number}), 
dependent parameters can be consolidated into a single head, trading 
parallelism for correctness.

\noindent\textbf{Fixed Head Count.} 
The 6-argument head configuration covers 95.2\% of functions in our benchmarks. 
APIs exceeding this limit can concatenate overflow parameters into the final 
head, preserving functionality with reduced parallelism.

\noindent\textbf{Training Cost.} 
Processing each head as an independent sequence incurs computational overhead 
compared to standard fine-tuning, acceptable for one-time model preparation 
but limiting rapid iteration.

\section{Future Work}
\label{sec:future_work}

\noindent\textbf{Open Source and Real-World Deployment.} 
We plan to release training code and model weights to facilitate adoption 
in latency-critical applications: robotic control systems requiring 
sub-100ms actions, real-time avatars, and interactive coding assistants.

\noindent\textbf{Vision-Language Extension.} 
Extending \method{} to VLMs would enable real-time Vision-Language-Action 
systems for embodied AI, with the key challenge being efficient visual 
token handling while maintaining parallel action decoding.

\noindent\textbf{On-Device Deployment.} 
Bridging server-side speedup to edge deployment requires optimization 
for mobile frameworks (LiteRT, Core ML, ExecuTorch) and adapting 
parallel decoding to device constraints.

\noindent\textbf{Adaptive Heads.} 
Dynamically adjusting active head count based on tool schemas at 
inference time would improve flexibility for diverse APIs.

\noindent\textbf{Scaling.} 
Investigating behavior on 30B+ models remains important---whether 
accuracy gains persist at scale or stronger baselines reduce the 
relative benefit.

\section{Conclusion}
\label{sec:conclusion}

We presented \method{}, a parallel decoding framework for real-time LLM function calling. 
Our key insight is that structured function outputs exhibit structural redundancy 
and weak causal dependencies among arguments---two properties that must be exploited jointly. 
We introduced special tokens that serve a dual role: compressing predictable tokens 
(4--6$\times$ reduction) while acting as mode selectors that enable independent 
generation of function name and arguments. 
This synergistic design allows parallel decoding streams to share the prefix KV cache 
with negligible overhead, achieving 3--6$\times$ end-to-end speedup 
while maintaining competitive accuracy across five benchmarks. 
Combined with quantization, \method{} enables 4B-scale models to achieve 
16+\,Hz real-time function calling on consumer GPUs, 
bridging the gap between LLM capabilities and latency-critical applications 
such as embodied AI, game AI, and interactive agents.
\section*{Impact Statement}

This paper advances real-time LLM function calling, enabling 
latency-critical applications such as embodied AI, game AI, 
and interactive agents. We foresee positive societal impacts 
including more responsive assistive robotics, accessible AI 
tutoring systems, and democratized deployment of capable AI 
on consumer hardware.

Potential risks include misuse in autonomous systems operating 
without adequate human oversight. We recommend that practitioners 
deploying \method{} in safety-critical domains implement 
appropriate safeguards, monitoring mechanisms, and human-in-the-loop 
controls. The authors have no conflicts of interest to disclose.

\bibliographystyle{icml2026}
\bibliography{references}

\newpage
\appendix
\onecolumn

\section{Other Related Work}
\label{app:related}

\paragraph{LLM Function Calling and Tool Use.}
The ability of LLMs to interact with external tools has been extensively studied. 
Toolformer \cite{schick2023toolformer} explored self-supervised tool learning, 
while Gorilla \cite{patil2024gorilla} and ToolLLM \cite{qin2024toolllm} improved 
accuracy through retrieval-augmented approaches. xLAM \cite{zhang2025xlam} 
addresses function calling through data augmentation strategies. 
At the orchestration level, several works explore parallelism and scheduling: 
LLM Compiler \cite{kim2024llm} investigates parallel function calling, 
Skeleton-of-Thought \cite{ning2024skeletonofthought} explores parallel generation 
via prompting strategies, AsyncLM \cite{gim2024asynchronous} proposes asynchronous 
execution patterns, and GhostShell \cite{gong2025ghostshell} addresses concurrent 
tool execution in embodied settings. LLMA \cite{yang2023inference} explores 
reference-based acceleration techniques.
For evaluation, BFCL \cite{patil2024gorilla} and StableToolBench \cite{guo2024stabletoolbench} 
provide standardized benchmarks, while Mobile Actions \cite{functiongemma2025} targets 
edge deployment scenarios. Google's FunctionGemma \cite{functiongemma2025} represents 
industry efforts toward compact, edge-deployable models.

\paragraph{Speculative and Parallel Decoding.}
Speculative decoding \cite{leviathan2023fast,chen2023accelerating} accelerates 
inference using draft-and-verify mechanisms. Medusa \cite{cai2024medusa} and 
the EAGLE series \cite{li2024eagle,li2024eagle2,li2025eagle3} introduce parallel 
decoding heads, while Hydra \cite{ankner2024hydra} explores sequentially-dependent 
draft heads. Multi-token prediction \cite{gloeckle2024better} trains models to 
predict multiple future tokens. A comprehensive survey is provided by 
\citet{xia2024unlocking}. Unlike these general-purpose methods, \method{} 
exploits the structural properties of function calling, achieving parallelism 
without additional parameters or draft models.

\paragraph{Structured Output Generation.}
SGLang \cite{zheng2024sglang} optimizes structured output generation through 
techniques such as jump-forward decoding for deterministic tokens. 
Grammar-constrained approaches include Outlines \cite{willard2023efficient} and 
XGrammar \cite{dong2024xgrammar}. While these methods improve output validity 
and efficiency, they remain fundamentally sequential. \method{} takes an 
orthogonal approach by reducing the output representation itself while enabling 
parallel generation across independent streams.

\paragraph{Efficient LLM Serving.}
Modern serving systems have improved LLM inference efficiency through various 
techniques. vLLM \cite{kwon2023efficient} introduced PagedAttention for KV cache 
management, while SGLang \cite{zheng2024sglang} proposed RadixAttention for 
prefix sharing. Quantization techniques such as AWQ \cite{lin2025awq} and GPTQ 
\cite{frantar2023gptq} enable deployment on resource-constrained devices. 
\method{} is compatible with these optimizations and can be combined with 
existing serving infrastructure.

\paragraph{Real-Time LLM Applications.}
Vision-language-action models such as ST-2 \cite{zitkovich2023rt}, OpenVLA 
\cite{kim2025openvla}, and $\pi_0$ \cite{black2024pizero} have demonstrated 
promising capabilities for robotic control. In game AI, Voyager \cite{wang2024voyager} 
and SIMA \cite{simateam2024scaling} explore open-world control capabilities. 
For GUI automation, AutoDroid \cite{wen2024autodroid} and Mobile-Agent 
\cite{wang2024mobile} explore LLM-powered device control. These application 
domains commonly face latency challenges that motivate our work on parallel decoding.

\section{Special Token Definitions}
\label{app:tokens}

\method{} introduces 17 special tokens organized into four 
functional groups. New tokens are appended to the vocabulary 
with IDs $|\mathcal{V}_{\text{base}}|$ through 
$|\mathcal{V}_{\text{base}}| + 16$. Embeddings are initialized 
by averaging semantically similar existing tokens.

\begin{table}[h]
\caption{Special token definitions and their functions.}
\label{tab:special_tokens}
\centering
\begin{tabular}{llc}
\toprule
\textbf{Group} & \textbf{Tokens} & \textbf{Count} \\
\midrule
Content Control & \texttt{<content>}, \texttt{</content>} & 2 \\
Function Control & \texttt{<function>}, \texttt{</function>} & 2 \\
Argument Control & \texttt{<arg$k$>}, \texttt{</arg$k$>} for $k \in \{1, \ldots, 6\}$ & 12 \\
Placeholder & \texttt{<|null|>} & 1 \\
\midrule
\textbf{Total} & & \textbf{17} \\
\bottomrule
\end{tabular}
\end{table}

\textbf{Example output format:}
\begin{verbatim}
<content>I'll check the weather for you.</content>
<function>get_weather</function>
<arg1>Beijing</arg1>
<arg2>2024-12-24</arg2>
<arg3>celsius</arg3>
<arg4><|null|></arg4>
<arg5><|null|></arg5>
<arg6><|null|></arg6>
\end{verbatim}

\section{Parameter Count Statistics}
\label{app:param_stats}

We analyze the distribution of function parameter counts across all 
evaluation benchmarks to justify our design choice of 6 parallel argument heads.

\subsection{Distribution Analysis}

Table~\ref{tab:param_count} presents the number of samples with more than 
6 parameters across all evaluation datasets (excluding SealTools 1--6 tools 
subsets, which are designed for controlled parameter count evaluation).

\begin{table}[h]
\centering
\caption{Parameter count distribution across benchmarks. 
Samples with $>$6 parameters require sequential handling for additional arguments.}
\label{tab:param_count}
\begin{tabular}{lrrrr}
\toprule
\textbf{Dataset} & \textbf{Total} & \textbf{$>$6 params} & \textbf{Ratio} & \textbf{Max Params} \\
\midrule
BFCL-v3 Exec Multiple & 50 & 0 & 0.00\% & 6 \\
BFCL-v3 Exec Simple & 100 & 0 & 0.00\% & 6 \\
BFCL-v3 Live Multiple & 1,053 & 167 & 15.86\% & 21 \\
BFCL-v3 Live Simple & 258 & 14 & 5.43\% & 10 \\
BFCL-v3 Multiple & 200 & 0 & 0.00\% & 6 \\
BFCL-v3 Simple & 400 & 0 & 0.00\% & 6 \\
SealTools (in-domain) & 199 & 4 & 2.01\% & 8 \\
SealTools (out-domain) & 94 & 0 & 0.00\% & 6 \\
Mobile Actions & 1,283 & 0 & 0.00\% & 4 \\
OpenFunctions & 109 & 0 & 0.00\% & 4 \\
ToolAlpaca (real) & 92 & 0 & 0.00\% & 6 \\
ToolAlpaca (simulated) & 51 & 0 & 0.00\% & 6 \\
\midrule
\textbf{Total} & \textbf{3,889} & \textbf{185} & \textbf{4.76\%} & -- \\
\bottomrule
\end{tabular}
\end{table}

\subsection{Conclusion}

Our default configuration of 6 parallel argument heads covers 95.2\% of all 
function calls in the evaluated benchmarks. The 4.76\% of cases with $>$6 
parameters are concentrated in BFCL-v3 Live datasets, which contain 
real-world APIs with extensive optional parameters (e.g., REST APIs with 
many query parameters). For these cases, our model generates the first 6 
arguments in parallel; additional arguments---typically optional---can be 
handled through sequential generation or omitted based on the tool schema's 
\texttt{required} field specification.

Notably, all synthetic and controlled benchmarks (BFCL Non-Live, SealTools, 
Mobile Actions, OpenFunctions, ToolAlpaca) have a maximum of 6 parameters, 
confirming that our head count aligns with common API design practices.

\section{Training Data Statistics}
\label{app:data}

\begin{table}[h]
\caption{Training data composition and statistics.}
\label{tab:data-composition}
\centering
\begin{tabular}{lcc}
\toprule
\textbf{Source} & \textbf{Samples} & \textbf{Percentage} \\
\midrule
xLAM-60K (filtered) & 48,000 & 40\% \\
Synthetic (Qwen3-235B generated) & 70,000 & 58\% \\
Special (pass-through functions) & 2,000 & 2\% \\
\midrule
\textbf{Total Groups} & \textbf{120,000} & \textbf{100\%} \\
\textbf{Total Samples ($\times$8 heads)} & \textbf{960,000} & -- \\
\bottomrule
\end{tabular}
\end{table}

\textbf{Argument distribution after augmentation:} 
1 arg: 18\%, 2 args: 20\%, 3 args: 18\%, 4 args: 17\%, 
5 args: 15\%, 6 args: 12\%. This balanced distribution 
addresses the natural imbalance where most real-world 
APIs use 1--2 arguments.

\section{Experimental Details}
\label{appendix:exp_details}

\subsection{Handling Parallel Tool Invocations}
\label{appendix:parallel_calls}

\subsubsection{Problem Statement}

Some benchmarks (e.g., Mobile Actions) contain samples with \textbf{parallel 
tool invocations}, where multiple functions are called simultaneously within 
a single turn. For example, a user request ``Turn on my flashlight and show 
me the nearest bookstore on the map'' expects two independent function calls:
\begin{verbatim}
tool_calls: [
  {"function": {"name": "turn_on_flashlight", "arguments": "{}"}},
  {"function": {"name": "show_map", "arguments": "{\"query\": \"nearest bookstore\"}"}}
]
\end{verbatim}

Since \method{} generates \textbf{one function call per inference pass}, 
we convert parallel invocations into sequential single-step evaluations 
using a history-based decomposition strategy.

\subsubsection{Conversion Strategy}

For $K$ parallel tool calls $\{c_1, c_2, ..., c_K\}$, we generate $K$ 
training/evaluation samples. Each sample $i$ receives:
\begin{itemize}
    \setlength{\itemsep}{0pt}
    \setlength{\parskip}{0pt}
    \item The original user query
    \item A \texttt{history} field containing preceding calls $\{c_1, ..., c_{i-1}\}$ 
    (empty for the first entry)
    \item Ground truth expecting only call $c_i$
\end{itemize}

\textbf{Key design choice for training:} During training data preparation, 
we apply \textbf{random shuffling} to the prefix calls for each version. 
This teaches the model that parallel calls have no inherent order dependency, 
which aligns with our parallel decoding assumption.

\subsubsection{Example}

\textbf{Original} (2 parallel calls: \texttt{turn\_on\_flashlight}, \texttt{show\_map}):

\textbf{Training Version 1} (target: \texttt{show\_map}):
\begin{verbatim}
User: "Turn on flashlight and show nearest bookstore"
History: [turn_on_flashlight()]
Assistant: -> show_map(query="nearest bookstore")  [TARGET]
\end{verbatim}

\textbf{Training Version 2} (target: \texttt{turn\_on\_flashlight}):
\begin{verbatim}
User: "Turn on flashlight and show nearest bookstore"
History: [show_map(query="nearest bookstore")]  [SHUFFLED]
Assistant: -> turn_on_flashlight()  [TARGET]
\end{verbatim}

\subsubsection{Implementation}

Algorithm~\ref{alg:parallel_conversion} illustrates the conversion process.

\begin{algorithm}[h]
\caption{Parallel Tool Call Decomposition}
\label{alg:parallel_conversion}
\begin{algorithmic}[1]
\REQUIRE Sample with query $q$, tool definitions $T$, parallel calls $C = \{c_1, ..., c_K\}$
\ENSURE Evaluation entries $E = \{e_1, ..., e_K\}$
\STATE $E \leftarrow \emptyset$
\FOR{$i = 1$ to $K$}
    \STATE $prefix \leftarrow \{c_1, ..., c_{i-1}\}$
    \STATE $prefix \leftarrow \text{Shuffle}(prefix)$ \COMMENT{Random order for training}
    \STATE $history \leftarrow \text{Format}(prefix)$
    \STATE $query_i \leftarrow \text{Concat}(environment, history, q)$
    \STATE $gt_i \leftarrow \text{CreateGroundTruth}(c_i, T)$
    \STATE $e_i \leftarrow (query_i, T, gt_i)$
    \STATE $E \leftarrow E \cup \{e_i\}$
\ENDFOR
\STATE \textbf{return} $E$
\end{algorithmic}
\end{algorithm}

\subsubsection{Benefits}

\begin{enumerate}
    \setlength{\itemsep}{0pt}
    \setlength{\parskip}{0pt}
    \item \textbf{Data augmentation}: $K$ parallel calls $\rightarrow$ $K$ training samples
    \item \textbf{Order invariance}: Random shuffling teaches the model that 
    parallel calls have no causal dependencies
    \item \textbf{Complete coverage}: Every function in the parallel set is 
    learned as a prediction target
    \item \textbf{Consistent evaluation}: Each decomposed entry is evaluated 
    independently; a parallel invocation is considered fully correct only if 
    all $K$ individual entries are correctly predicted
\end{enumerate}

\subsection{Parameter Normalization}

To ensure consistent evaluation across all benchmarks, we normalize 
tool definitions to have at most 6 parameters per function. 
Parameters are ordered as follows:
\begin{enumerate}
    \setlength{\itemsep}{0pt}
    \setlength{\parskip}{0pt}
    \item Required parameters in their original specification order
    \item Optional parameters in alphabetical order
\end{enumerate}

This normalization aligns with \method{}'s 6-argument head design 
and ensures that ground truth parameter ordering matches the 
tool definitions provided to the model.

\subsection{Latency Measurement Methodology}
\label{appendix:latency_method}

We measure inference latency using consistent methodology across 
all models and frameworks to ensure fair comparison.

\textbf{Measurement Protocol.}
\begin{itemize}
    \setlength{\itemsep}{0pt}
    \setlength{\parskip}{0pt}
    \item \textbf{Metric}: End-to-end wall-clock time from request 
    submission to completion, including all framework overhead.
    \item \textbf{Warmup}: 5 requests discarded before measurement 
    to ensure stable GPU state and cache initialization.
    \item \textbf{Sample size}: Full test split (1,283 samples for 
    Mobile Actions) to capture latency distribution.
    \item \textbf{Statistics}: P50 (median) and P90 reported; 
    P50 reflects typical performance while P90 captures tail latency.
\end{itemize}

\textbf{Framework Configuration.} All vLLM experiments use:
\begin{itemize}
    \setlength{\itemsep}{0pt}
    \setlength{\parskip}{0pt}
    \item \texttt{enable\_prefix\_caching=True}
    \item \texttt{gpu\_memory\_utilization=0.85}
    \item \texttt{tensor\_parallel\_size=1}
    \item Single request per batch (serial mode) to measure 
    per-request latency rather than throughput
\end{itemize}

\textbf{Stop Token Configuration.} For \method{} models, we configure 
both string-based and token-ID-based stopping to ensure immediate 
termination on special tokens:
\begin{verbatim}
stop = ["<|null|>", "</function>", "</arg1>", ..., "<|im_end|>"]
stop_token_ids = [tokenizer.encode(s)[-1] for s in stop]
\end{verbatim}
This dual configuration is critical for achieving optimal latency, 
as token-ID-based stopping triggers immediately upon generation 
without the 1-step delay inherent in string matching.

\subsection{Mobile Actions Latency Benchmark Details}
\label{appendix:mobile_latency}

Table~\ref{tab:mobile_latency_full} provides comprehensive latency 
statistics for the Mobile Actions benchmark discussed in 
Section~\ref{sec:exp_mobile}.

\begin{table}[h]
\centering
\small
\caption{Full latency statistics on Mobile Actions (1,283 samples, RTX 4090).}
\label{tab:mobile_latency_full}
\begin{tabular}{lcccccc}
\toprule
\textbf{Model} & \textbf{P50 (ms)} & \textbf{P90 (ms)} & \textbf{P95 (ms)} & \textbf{P99 (ms)} & \textbf{Mean (ms)} & \textbf{Avg Tok} \\
\midrule
Qwen3-4B (Baseline) & 468.5 & 667.0 & 731.2 & 845.3 & 489.7 & 39.2 \\
Qwen2.5-0.5B (Baseline) & 110.1 & 154.0 & 166.2 & 188.6 & 97.7 & 40.2 \\
FunctionGemma (270M) & 61.1 & 139.5 & 149.4 & 173.7 & 71.7 & 30.3 \\
\midrule
\textbf{ST-Qwen-0.5B (Ours)} & \textbf{51.0} & \textbf{74.5} & \textbf{82.2} & \textbf{90.0} & \textbf{50.1} & $\sim$8 \\
\bottomrule
\end{tabular}
\end{table}

\textbf{Key Observations.}
\begin{itemize}
    \setlength{\itemsep}{0pt}
    \setlength{\parskip}{0pt}
    \item ST-Qwen-0.5B achieves the lowest latency across all percentiles.
    \item The P90/P50 ratio for ST-Qwen-0.5B (1.27) is significantly 
    lower than FunctionGemma (2.27), indicating more consistent performance.
    \item Average output tokens for ST-Qwen-0.5B ($\sim$8) are 
    3--5$\times$ fewer than baselines (30--40), directly contributing 
    to reduced decode time.
\end{itemize}

\subsection{Hardware Specifications}

All experiments were conducted on the following hardware:

\begin{table}[h]
\centering
\small
\caption{Hardware configurations used in experiments.}
\label{tab:hardware}
\begin{tabular}{ll}
\toprule
\textbf{Component} & \textbf{Specification} \\
\midrule
GPU  & NVIDIA RTX 4090 (24GB VRAM) \\
GPU  & NVIDIA H100/H200 (80GB/141GB VRAM) \\
CUDA & 12.9 \\
PyTorch & 2.8.0 \\
vLLM & 0.12.0 \\
\bottomrule
\end{tabular}
\end{table}

Training experiments were conducted on H100/H200 GPUs. ST-Qwen-0.5B converges 
in approximately 3--4 H100-hours for initial training, and $\sim$1 GPU-hour 
for domain adaptation on Mobile Actions.

\section{Training Hyperparameters}
\label{app:hyperparameters}

Table~\ref{tab:full-hyperparameters} presents the complete training configuration used for \method{} models. We highlight several key design choices:

\textbf{LoRA Configuration.} We target MLP layers (gate\_proj, up\_proj, down\_proj) rather than attention layers, as we find this provides greater capacity for learning the distinct output modes required by parallel heads. The relatively high rank ($r=512$) is necessary to capture diverse head-specific patterns without interference (see ablation in Section~\ref{sec:ablation_rank}).

\textbf{Head Loss Weighting.} We apply higher weights to the function head ($w_1=2.0$) since correct function selection is critical---an incorrect function name invalidates the entire output regardless of argument quality. Argument heads receive progressively increasing weights ($w_2$ through $w_7$) to counteract the natural data imbalance where higher-indexed arguments appear less frequently.

\textbf{Focal Weights.} For argument heads, we apply token-level focal weights that increase with head index. This addresses the challenge that later argument slots (arg4--arg6) have fewer training examples due to the skewed distribution of argument counts in real-world APIs.

\begin{table}[t]
\caption{Complete training hyperparameters for \method{} in ST-Qwen3-4B}
\label{tab:full-hyperparameters}
\centering
\begin{tabular}{ll}
\toprule
\textbf{Parameter} & \textbf{Value} \\
\midrule
\multicolumn{2}{l}{\textit{LoRA Configuration}} \\
Rank ($r$) & 512 \\
Alpha ($\alpha$) & 1024 \\
Dropout & 0.05 \\
Target Modules & gate\_proj, up\_proj, down\_proj \\
\midrule
\multicolumn{2}{l}{\textit{Learning Rates}} \\
Embedding & $1 \times 10^{-6}$ \\
LoRA & $1 \times 10^{-5}$ \\
LM Head & $1 \times 10^{-6}$ \\
Scheduler & Cosine with warmup \\
Warmup Ratio & 0.02 \\
\midrule
\multicolumn{2}{l}{\textit{Training}} \\
Sequence Length & 2048 tokens \\
Batch Size (per device) & 8 groups \\
Gradient Accumulation & 4 steps \\
Effective Batch Size & 32 groups (256 samples) \\
Epochs & 3--4 \\
Precision & bfloat16 \\
Flash Attention & Enabled (v2) \\
\midrule
\multicolumn{2}{l}{\textit{Head Loss Weights ($w_h$)}} \\
$w_0$ (content) & 0.01 \\
$w_1$ (function) & 2.0 \\
$w_{2..7}$ (arg1--arg6) & 1.1, 1.2, 1.3, 1.4, 1.5, 1.6 \\
\midrule
\multicolumn{2}{l}{\textit{Token-Level Focal Weights ($\gamma$, non-null only)}} \\
$\gamma_{2,3}$ (arg1--arg2) & 1.0 \\
$\gamma_4$ (arg3) & 1.5 \\
$\gamma_5$ (arg4) & 2.0 \\
$\gamma_6$ (arg5) & 2.5 \\
$\gamma_7$ (arg6) & 3.0 \\
\bottomrule
\end{tabular}
\end{table}

\section{Batch Scaling Analysis}
\label{app:batch_scaling}

This appendix validates \method{}'s core assumption that parallel 
decoding incurs negligible overhead due to the memory-bandwidth-bound 
nature of autoregressive decoding.

\subsection{Experimental Setup}

We measure per-token decode latency across varying batch sizes 
(1, 2, 4, 8, 16, 32, 64, 128) for six model configurations: 
Qwen2.5-0.5B, 1.5B, 3B, 7B, 14B, and Qwen3-4B.
All experiments use identical input prompts and measure 
wall-clock time for the decode phase only, excluding prefill.
We define \textbf{efficiency} as:
\begin{equation}
    \text{Efficiency}(B) = \frac{T_d^{(B=1)}}{T_d^{(B)}}
\end{equation}
where $T_d^{(B)}$ is the per-token decode latency at batch size $B$.
An efficiency of 100\% indicates perfect scaling (no overhead from batching); 
lower values indicate compute saturation.

\subsection{Results}

Table~\ref{tab:batch_efficiency} presents efficiency across batch sizes.
The key finding is that \textbf{8-head parallel decoding 
(corresponding to B=8) achieves 93.0\% average efficiency}, 
with per-token overhead of only +8.2\% compared to single-sequence decoding.

\begin{table}[h]
\centering
\caption{Batch scaling efficiency (\%) across model sizes. 
Higher values indicate better utilization of idle compute capacity.
\method{} uses 8 heads, corresponding to B=8.}
\label{tab:batch_efficiency}
\small
\begin{tabular}{lcccccccc}
\toprule
\textbf{Model} & \textbf{B=1} & \textbf{B=2} & \textbf{B=4} & \textbf{B=8} & \textbf{B=16} & \textbf{B=32} & \textbf{B=64} & \textbf{B=128} \\
\midrule
Qwen2.5-0.5B & 100.0 & 96.4 & 95.6 & 90.6 & 81.4 & 67.3 & 52.8 & 37.1 \\
Qwen2.5-1.5B & 100.0 & 98.3 & 97.5 & 92.7 & 86.5 & 78.7 & 64.7 & 47.1 \\
Qwen2.5-3B & 100.0 & 99.1 & 98.4 & 93.5 & 89.8 & 83.5 & 72.1 & 56.2 \\
Qwen3-4B & 100.0 & 98.7 & 96.3 & 92.5 & 83.7 & 74.4 & 58.1 & 45.8 \\
Qwen2.5-7B & 100.0 & 87.2 & 91.5 & 91.5 & 90.9 & 84.4 & 75.0 & 62.0 \\
Qwen2.5-14B & 100.0 & 99.3 & 98.4 & 97.4 & 95.2 & 90.6 & 78.1 & 66.5 \\
\midrule
\textbf{Average} & 100.0 & 96.5 & 96.3 & \textbf{93.0} & 87.9 & 79.8 & 66.8 & 52.5 \\
\bottomrule
\end{tabular}
\end{table}

Table~\ref{tab:batch_overhead} presents the same data as wall-clock 
overhead relative to B=1.

\begin{table}[h]
\centering
\caption{Wall-clock overhead (\%) relative to single-sequence decoding (B=1).
Lower values indicate less overhead from parallel execution.}
\label{tab:batch_overhead}
\small
\begin{tabular}{lcccccccc}
\toprule
\textbf{Model} & \textbf{B=1} & \textbf{B=2} & \textbf{B=4} & \textbf{B=8} & \textbf{B=16} & \textbf{B=32} & \textbf{B=64} & \textbf{B=128} \\
\midrule
Qwen2.5-0.5B & +0.0 & +3.7 & +4.6 & +10.3 & +19.6 & +34.5 & +62.7 & +125.3 \\
Qwen2.5-1.5B & +0.0 & +1.7 & +2.6 & +7.6 & +15.6 & +23.9 & +44.4 & +93.2 \\
Qwen2.5-3B & +0.0 & +0.9 & +1.6 & +6.9 & +10.0 & +19.1 & +37.0 & +76.9 \\
Qwen3-4B & +0.0 & +1.3 & +3.9 & +5.5 & +10.4 & +23.2 & +37.6 & +68.0 \\
Qwen2.5-7B & +0.0 & +12.8 & +15.7 & +16.3 & +21.6 & +26.6 & +41.1 & +68.7 \\
Qwen2.5-14B & +0.0 & +0.8 & +1.7 & +2.7 & +4.7 & +10.1 & +23.2 & +44.9 \\
\midrule
\textbf{Average} & +0.0 & +3.5 & +5.0 & \textbf{+8.2} & +13.7 & +22.9 & +41.0 & +79.5 \\
\bottomrule
\end{tabular}
\end{table}

\subsection{Analysis}

\textbf{Memory-Bound Regime.} 
The high efficiency at B=8 confirms that autoregressive decoding 
operates in a memory-bandwidth-bound regime for typical model sizes. 
The GPU's compute units remain underutilized during single-sequence 
decoding, allowing additional sequences to be processed with 
minimal latency increase.

\textbf{Efficiency Inflection Point.} 
Efficiency drops below 80\% at B$\geq$32 for most models, 
indicating the transition from memory-bound to compute-bound regime.
\method{}'s 8 parallel heads remain well within the ``comfort zone'' 
where parallelization overhead is negligible.

\textbf{Model Size Effect.} 
Larger models (7B, 14B) maintain higher efficiency at larger batch sizes, 
as their higher memory bandwidth requirements keep the system 
memory-bound for longer. This suggests \method{}'s parallel decoding 
strategy scales favorably with model size.

\textbf{Implications for \method{}.} 
With 8 parallel heads and +8.2\% average overhead, the scheduling 
term $T_o(H)$ in Equation~\ref{eq:latency} is effectively absorbed 
into a small multiplicative factor on $T_d$:
\begin{equation}
    T_{\text{ours}} \approx T_p + \max_{i}(N_i) \cdot (1.08 \cdot T_d)
\end{equation}
Combined with the 4--6$\times$ token compression (Table~\ref{tab:token_compression}), 
this yields the 3--6$\times$ end-to-end speedup observed in Section~\ref{sec:exp_speedup}.

\section{Token Compression Details}
\label{app:compression_details}

This appendix provides detailed token compression statistics 
supporting the analysis in Section~\ref{sec:parallel_decoding}.

\subsection{Methodology}

We measure token compression by comparing output lengths between 
baseline models (generating full JSON-formatted function calls) and 
\method{} models (generating only value tokens across parallel heads).
For \method{}, we report the \textbf{bottleneck head} token count---the 
maximum across all parallel heads---as this determines end-to-end latency 
per Equation~\ref{eq:latency}.

Statistics are computed exclusively on samples where \textbf{both} 
baseline and \method{} models produce correct outputs, ensuring 
fair comparison on equivalent task difficulty.
The compression ratio (CR) is defined as:
\begin{equation}
    \text{CR} = \frac{\text{Baseline tokens}}{\text{ST bottleneck-head tokens}}
\end{equation}

\subsection{Overall Results}

Table~\ref{tab:token_compression_detailed} presents detailed statistics 
across model sizes, including mean, median (P50), and 90th percentile (P90) 
for both baseline and \method{} token counts.

\begin{table*}[t]
\centering
\caption{Detailed token compression analysis. BL = Baseline tokens, ST = ST bottleneck-head tokens, CR = Compression Ratio. All statistics computed on samples where both Baseline and ST models produce correct outputs.}
\label{tab:token_compression_detailed}
\begin{tabular}{l|rrr|rrr|rrr|r}
\toprule
 & \multicolumn{3}{c|}{\textbf{Mean}} & \multicolumn{3}{c|}{\textbf{P50 (Median)}} & \multicolumn{3}{c|}{\textbf{P90}} & \\
\textbf{Model} & BL & ST & CR & BL & ST & CR & BL & ST & CR & \textbf{N} \\
\midrule
Qwen2.5-0.5B & 44.1 & 8.7 & 5.06$\times$ & 36.0 & 6.0 & 6.00$\times$ & 69.0 & 17.0 & 4.06$\times$ & 3328 \\
Qwen2.5-1.5B & 38.3 & 8.9 & 4.32$\times$ & 33.0 & 7.0 & 4.71$\times$ & 64.0 & 20.0 & 3.20$\times$ & 3590 \\
Qwen2.5-3B & 39.2 & 8.9 & 4.38$\times$ & 33.0 & 7.0 & 4.71$\times$ & 64.0 & 20.0 & 3.20$\times$ & 3939 \\
Qwen3-4B$^\dagger$ & 38.4 & 8.9 & 4.32$\times$ & 34.0 & 7.0 & 4.86$\times$ & 65.0 & 19.0 & 3.42$\times$ & 4074 \\
Qwen2.5-7B & 48.3 & 9.0 & 5.35$\times$ & 35.0 & 7.0 & 5.00$\times$ & 68.0 & 20.0 & 3.40$\times$ & 4094 \\
Qwen2.5-14B & 40.2 & 8.9 & 4.51$\times$ & 34.0 & 7.0 & 4.86$\times$ & 66.0 & 19.0 & 3.47$\times$ & 3946 \\
\bottomrule
\end{tabular}
\vspace{1mm}
\\ \footnotesize{$\dagger$: Qwen3 backbone. N = number of both-correct samples.}
\end{table*}

Key observations:
\begin{itemize}
    \setlength{\itemsep}{0pt}
    \setlength{\parskip}{0pt}
    \item \textbf{Consistent compression across scales}: All models achieve 
    4.3--5.4$\times$ mean compression, indicating that the structural 
    redundancy in function call outputs is model-agnostic.
    \item \textbf{Higher median than mean}: P50 compression (4.7--6.0$\times$) 
    exceeds mean compression, suggesting that outliers with longer outputs 
    (complex nested arguments) pull down the average.
    \item \textbf{Stable RT token counts}: \method{} bottleneck-head tokens 
    remain remarkably stable (8.7--9.0 mean) across model sizes, 
    reflecting the consistent information content of function calls 
    regardless of baseline verbosity.
\end{itemize}

\subsection{Per-Benchmark Breakdown}

Table~\ref{tab:token_compression_by_group} presents compression ratios 
by benchmark group, revealing how compression varies across task types.

\begin{table*}[t]
\centering
\caption{Token compression ratio by benchmark group. CR = Compression Ratio (Baseline / RT). Results on both-correct samples. BFCL-NL = BFCL Non-Live, BFCL-L = BFCL Live, BFCL-E = BFCL Exec.}
\label{tab:token_compression_by_group}
\begin{tabular}{l|ccccccc|c}
\toprule
\textbf{Model} & \textbf{BFCL-NL} & \textbf{BFCL-L} & \textbf{BFCL-E} & \textbf{SealTools} & \textbf{MobileAct} & \textbf{OpenFunc} & \textbf{ToolAlpaca} & \textbf{Overall} \\
\midrule
Qwen2.5-0.5B & 7.05$\times$ & 5.52$\times$ & 4.53$\times$ & 5.16$\times$ & 4.17$\times$ & 7.91$\times$ & 5.14$\times$ & 5.06$\times$ \\
Qwen2.5-1.5B & 5.67$\times$ & 5.00$\times$ & 3.85$\times$ & 4.51$\times$ & 3.38$\times$ & 5.19$\times$ & 6.04$\times$ & 4.32$\times$ \\
Qwen2.5-3B & 5.73$\times$ & 4.83$\times$ & 3.55$\times$ & 4.53$\times$ & 3.73$\times$ & 5.20$\times$ & 4.48$\times$ & 4.38$\times$ \\
Qwen3-4B$^\dagger$ & 5.71$\times$ & 4.89$\times$ & 3.16$\times$ & 4.55$\times$ & 3.54$\times$ & 5.17$\times$ & 4.47$\times$ & 4.32$\times$ \\
Qwen2.5-7B & 6.77$\times$ & 6.43$\times$ & 3.91$\times$ & 5.97$\times$ & 3.46$\times$ & 5.29$\times$ & 20.54$\times$ & 5.35$\times$ \\
Qwen2.5-14B & 5.81$\times$ & 5.44$\times$ & 3.38$\times$ & 4.91$\times$ & 3.31$\times$ & 5.17$\times$ & 6.16$\times$ & 4.51$\times$ \\
\bottomrule
\end{tabular}
\vspace{1mm}
\\ \footnotesize{$\dagger$: Qwen3 backbone.}
\end{table*}

Key observations:
\begin{itemize}
    \setlength{\itemsep}{0pt}
    \setlength{\parskip}{0pt}
    \item \textbf{Highest compression on simple APIs}: BFCL Non-Live and 
    OpenFunc achieve 5.7--7.9$\times$ compression, as these benchmarks 
    feature functions with fewer, simpler arguments.
    \item \textbf{Lower compression on execution tasks}: BFCL Exec and 
    MobileAct show 3.2--4.5$\times$ compression, reflecting more complex 
    argument structures (e.g., nested objects, longer string values).
    \item \textbf{Outlier in ToolAlpaca}: Qwen2.5-7B achieves 20.54$\times$ 
    compression on ToolAlpaca, likely due to baseline verbosity on 
    specific samples; this outlier does not significantly affect 
    overall conclusions.
\end{itemize}

\subsection{Relationship to End-to-End Speedup}

The token compression ratio provides a \textbf{theoretical upper bound} 
on decode-phase speedup. The actual end-to-end speedup 
(3--6$\times$, Section~\ref{sec:exp_speedup}) is lower due to:
\begin{enumerate}
    \setlength{\itemsep}{0pt}
    \setlength{\parskip}{0pt}
    \item \textbf{Prefill overhead}: $T_p$ is identical for both methods 
    and does not benefit from token compression.
    \item \textbf{Parallelization overhead}: The +8.2\% overhead from 
    8-head parallel decoding (Appendix~\ref{app:batch_scaling}).
    \item \textbf{Framework overhead}: Scheduling, memory allocation, 
    and other system-level costs in inference frameworks.
\end{enumerate}

Despite these factors, \method{} achieves 70--80\% of the theoretical 
compression-based speedup in practice, demonstrating efficient 
translation of token reduction into latency improvement.

\section{Limitations: Parameter Independence Assumption}
\label{app:limitations}

\subsection{The Independence Assumption}

Our parallel decoding assumes that function arguments are 
\textbf{conditionally independent} given the user query and function name:
\begin{equation}
    P(\text{arg}_1, ..., \text{arg}_k | q, f) \approx \prod_{i=1}^{k} P(\text{arg}_i | q, f)
\end{equation}

This assumption is justified by the design philosophy of JSON-based 
function calling: parameters are defined as unordered key-value pairs, 
and well-designed APIs treat each parameter as capturing an independent 
aspect of the request.

\subsection{Empirical Validation}

Our benchmark results provide strong empirical support for this assumption:

\begin{itemize}
    \setlength{\itemsep}{0pt}
    \setlength{\parskip}{0pt}
    \item \textbf{Accuracy improvement}: \method{} achieves equal or higher 
    accuracy than sequential baselines across all benchmarks 
    (Table~\ref{tab:main_results}). If parameter dependencies were 
    prevalent and important, removing sequential conditioning should 
    \textit{decrease} accuracy---the opposite of what we observe.
    \item \textbf{Order invariance in training}: Our parallel-to-multi-turn 
    conversion (Appendix~\ref{appendix:parallel_calls}) shuffles parameter 
    order during training. The model's strong performance indicates it 
    learns to predict parameters independently of their presentation order.
\end{itemize}

\subsection{Theoretical Edge Cases}

We acknowledge that some API designs could introduce parameter dependencies:

\textbf{Example 1: Runtime-dependent validation}
\begin{verbatim}
read_file_range(file_path, start_line, end_line)
\end{verbatim}
Here, \texttt{end\_line}'s validity depends on the actual file length. 
However, this is a \textit{runtime} constraint, not a \textit{generation} 
dependency---a sequential model would also generate \texttt{end\_line=100} 
if the user requests it, regardless of whether the file has 100 lines.

\textbf{Example 2: Cascading inference}
\begin{verbatim}
format_convert(input_string, detected_format, output_format)
\end{verbatim}
If \texttt{detected\_format} must be inferred from \texttt{input\_string}, 
this creates a true dependency. However:
\begin{enumerate}
    \setlength{\itemsep}{0pt}
    \setlength{\parskip}{0pt}
    \item This is poor API design---\texttt{detected\_format} should be 
    computed internally by the function, not passed as a parameter.
    \item In practice, both values can often be inferred independently 
    from context (e.g., ``convert 2024-01-15 to MM/DD/YYYY'' allows 
    inferring both the input format and the input value).
\end{enumerate}

\subsection{Handling True Dependencies}

For the rare cases where genuine parameter dependencies exist, 
\method{} supports two fallback strategies:

\begin{enumerate}
    \setlength{\itemsep}{0pt}
    \setlength{\parskip}{0pt}
    \item \textbf{Consolidation}: Dependent parameters can be combined 
    into a single head as a structured value (e.g., a list or nested object).
    \item \textbf{Multi-turn refinement}: For complex cases, the model can 
    generate a partial call first, then refine in subsequent turns based 
    on execution feedback.
\end{enumerate}

\subsection{Conclusion}

The parameter independence assumption is both theoretically grounded 
(JSON's key-value design) and empirically validated (accuracy improvements). 
While edge cases exist, they represent either poor API design or runtime 
constraints that affect sequential models equally. Our 95.2\% benchmark 
coverage with 6 heads (Appendix~\ref{app:param_stats}) further confirms 
that real-world APIs align with this assumption.

\section{Compatibility with Speculative Decoding}
\label{app:speculative}

As discussed in Section~\ref{sec:intro}, speculative decoding methods 
are orthogonal to \method{}: they accelerate per-token generation 
through draft-and-verify mechanisms, while we reduce the total number 
of tokens to generate. This appendix validates that these approaches 
can be combined for additional benefits.

\subsection{Experimental Setup}

We evaluate speculative decoding on \method{}-trained models using 
the standard draft-model approach \cite{leviathan2023fast}. 
The target model is ST-Qwen2.5-14B, with ST-Qwen2.5-0.5B and 
ST-Qwen2.5-1.5B serving as draft models. All models share the same 
special token vocabulary and output format, enabling direct speculation 
without format conversion.

We test speculation depths $N \in \{2, 3, 4\}$ (number of tokens 
drafted per step) across all BFCL-v3 subsets and report forward pass 
reduction (vanilla forwards / speculative forwards) and token 
acceptance rate (accepted tokens / drafted tokens).

\subsection{Results}

Table~\ref{tab:speculative_summary} summarizes the results across 
draft model sizes and speculation depths.

\begin{table}[h]
\centering
\small
\caption{Speculative decoding results with ST-Qwen2.5-14B as target. 
Speedup = forward pass reduction (higher is better). 
Accept Rate = proportion of drafted tokens accepted by target model.}
\label{tab:speculative_summary}
\begin{tabular}{llcc}
\toprule
\textbf{Draft Model} & \textbf{N} & \textbf{Avg Speedup} & \textbf{Avg Accept Rate} \\
\midrule
\multirow{3}{*}{ST-Qwen-0.5B} 
  & 2 & 2.35$\times$ & 95.0\% \\
  & 3 & 2.83$\times$ & 94.0\% \\
  & 4 & 3.22$\times$ & 93.6\% \\
\midrule
\multirow{3}{*}{ST-Qwen-1.5B} 
  & 2 & 2.37$\times$ & 96.1\% \\
  & 3 & 2.87$\times$ & 95.5\% \\
  & 4 & 3.24$\times$ & 95.1\% \\
\bottomrule
\end{tabular}
\end{table}

Table~\ref{tab:speculative_detail} provides per-dataset breakdown 
for the best-performing configuration (ST-Qwen-1.5B draft, $N=4$).

\begin{table}[h]
\centering
\small
\caption{Per-dataset speculative decoding results (ST-Qwen-1.5B draft, $N=4$).}
\label{tab:speculative_detail}
\begin{tabular}{lrrrcc}
\toprule
\textbf{Dataset} & \textbf{Samples} & \textbf{Vanilla Fwds} & \textbf{Spec Fwds} & \textbf{Speedup} & \textbf{Accept Rate} \\
\midrule
BFCL-v3 Simple & 400 & 6,578 & 2,031 & 3.24$\times$ & 95.3\% \\
BFCL-v3 Live Simple & 244 & 4,698 & 1,401 & 3.35$\times$ & 93.0\% \\
BFCL-v3 Multiple & 200 & 3,333 & 1,096 & 3.04$\times$ & 96.1\% \\
BFCL-v3 Live Multiple & 886 & 17,196 & 6,593 & 2.61$\times$ & 90.0\% \\
BFCL-v3 Exec Simple & 100 & 1,844 & 489 & 3.77$\times$ & 98.2\% \\
BFCL-v3 Exec Multiple & 50 & 986 & 285 & 3.46$\times$ & 98.1\% \\
\midrule
\textbf{Total/Average} & 1,880 & 34,635 & 11,895 & \textbf{3.24$\times$} & \textbf{95.1\%} \\
\bottomrule
\end{tabular}
\end{table}

\subsection{Analysis}

\textbf{High acceptance rates.} Both draft models achieve $>$93\% 
token acceptance rates across all configurations, significantly 
higher than typical speculative decoding scenarios on general text 
generation (70--85\%). This is because \method{}'s simplified output 
format---consisting primarily of function names and argument values---exhibits 
lower entropy and higher predictability than free-form text, 
making smaller models more effective as drafters.

\textbf{Consistent speedup across datasets.} Forward pass reduction 
ranges from 2.6$\times$ to 3.8$\times$ depending on dataset complexity. 
Simpler datasets (Exec Simple/Multiple) achieve higher speedup due to 
more predictable outputs, while Live Multiple shows lower speedup 
due to greater output diversity.

\textbf{Draft model size trade-off.} The 1.5B draft model achieves 
slightly higher acceptance rates (+1--2\%) than the 0.5B model, 
translating to marginally better speedup. However, the 0.5B model 
may be preferable for memory-constrained deployments where loading 
a larger draft model is impractical.

\textbf{Speculation depth trade-off.} Increasing $N$ from 2 to 4 
improves speedup from $\sim$2.4$\times$ to $\sim$3.2$\times$, 
with only minor degradation in acceptance rate ($\sim$1\%). 
This suggests $N=4$ as a practical default for \method{} models.

\subsection{Combined Speedup Potential}

The total theoretical speedup when combining \method{} with 
speculative decoding is multiplicative:
\begin{equation}
    \text{Speedup}_{\text{combined}} = \text{CR} \times \text{Speedup}_{\text{spec}}
\end{equation}
where CR is the token compression ratio from \method{} 
(4--6$\times$, Table~\ref{tab:token_compression}) and 
$\text{Speedup}_{\text{spec}}$ is the forward pass reduction 
from speculative decoding (2--3$\times$).

For ST-Qwen2.5-14B with CR $\approx$ 4.5$\times$ 
(Table~\ref{tab:token_compression}) and 1.5B draft model at $N=4$ 
(3.24$\times$ forward reduction), the combined theoretical speedup 
reaches approximately \textbf{14.6$\times$} in forward passes 
compared to vanilla autoregressive decoding of the baseline 14B model.

\textbf{Practical considerations.} While forward pass reduction 
provides a useful proxy for latency improvement, actual wall-clock 
speedup depends on additional factors: draft model inference overhead, 
verification batch size, and memory bandwidth utilization. 
For memory-bound inference on consumer GPUs, \method{}'s token 
compression alone often saturates available bandwidth, making 
speculative decoding most beneficial for larger models or 
datacenter deployments where compute becomes the bottleneck. 
We leave detailed wall-clock benchmarking of combined approaches 
to future work.

\subsection{Conclusion}

These results confirm that \method{} and speculative decoding are 
complementary acceleration strategies operating on different axes: 
\method{} reduces \textit{what} to generate (fewer tokens through 
compression and parallelization), while speculative decoding 
accelerates \textit{how} to generate (fewer sequential forward 
passes through draft-and-verify). The exceptionally high acceptance 
rates ($>$93\%) suggest that \method{}'s simplified output format 
is particularly amenable to speculative approaches, opening 
promising directions for further optimization.

\end{document}